\documentclass[sigconf]{acmart}

\copyrightyear{2021}
\acmYear{2021}
\setcopyright{acmcopyright}\acmConference[SIGIR '21]{Proceedings of the 44th International ACM SIGIR Conference on Research and Development in Information Retrieval}{July 11--15, 2021}{Virtual Event, Canada}
\acmBooktitle{Proceedings of the 44th International ACM SIGIR Conference on Research and Development in Information Retrieval (SIGIR '21), July 11--15, 2021, Virtual Event, Canada}
\acmPrice{15.00}
\acmDOI{10.1145/3404835.3462959}
\acmISBN{978-1-4503-8037-9/21/07}

\usepackage{graphicx}
\usepackage{comment}
\usepackage{amsmath}
\usepackage{color}
\usepackage{wrapfig}
\usepackage{lipsum}
\usepackage{multirow}
\usepackage[caption=false]{subfig}
\usepackage{diagbox}
\usepackage{booktabs,bm}
\usepackage{color}
\usepackage{url}
\usepackage{slashbox}
\usepackage{comment}
\usepackage{makecell}
\usepackage{textcomp}
\usepackage{enumitem}
\usepackage{tikz} 
\usepackage{pgfplots}
\pgfplotsset{compat=newest}
\usepackage{pgf-pie}

\newcommand{\hide}[1]{} %
\usepackage{xspace}
\newcommand*{\eg}{\emph{e.g.},\@\xspace}
\newcommand*{\ie}{\emph{i.e.},\@\xspace}

\def\etal{\emph{et al.}\@\xspace}

\begin{document}
\fancyhead{}

\title{DeepQAMVS: Query-Aware Hierarchical Pointer Networks for Multi-Video Summarization}

\author{Safa Messaoud}
\affiliation{%
  \institution{University of Illinois at Urbana-Champaign}
  \city{}
  \state{}
  \country{}
}
\email{messaou2@illinois.edu}

\author{Ismini Lourentzou}
\affiliation{
  \institution{Virginia Tech}
  \city{}
  \state{}
  \country{}
}
\email{ilourentzou@vt.edu}

\author{Assma Boughoula}
\authornote{Both authors contributed equally.}
\affiliation{%
  \institution{University of Illinois at Urbana-Champaign}
  \city{}
  \state{}
  \country{}
}
\email{boughou1@illinois.edu}

\author{Mona Zehni}
\authornotemark[1]
\affiliation{%
  \institution{University of Illinois at Urbana-Champaign}
  \city{}
  \state{}
  \country{}
}
\email{mzehni2@illinois.edu}

\author{Zhizhen Zhao}
\affiliation{%
  \institution{University of Illinois at Urbana-Champaign}
  \city{}
  \state{}
  \country{}
}
\email{zhizhenz@illinois.edu}

\author{Chengxiang Zhai}
\affiliation{%
  \institution{University of Illinois at Urbana-Champaign}
  \city{}
  \state{}
  \country{}
}
\email{czhai@illinois.edu}

\author{Alexander G. Schwing}
\affiliation{%
  \institution{University of Illinois at Urbana-Champaign}
  \city{}
  \state{}
  \country{}
}
\email{aschwing@illinois.edu}

\begin{abstract}
The recent growth of web video sharing platforms has increased the demand for systems that can efficiently browse, retrieve and summarize video content. Query-aware multi-video summarization is a promising technique that caters to this demand. 
In this work, we introduce a novel Query-Aware Hierarchical Pointer Network for Multi-Video Summarization, termed DeepQAMVS, that jointly optimizes multiple criteria: (1) conciseness, (2) representativeness of important query-relevant events and (3) chronological soundness. We design a hierarchical attention model that factorizes over three distributions, each collecting evidence from a different modality, followed by a pointer network that selects frames to include in the summary. 
DeepQAMVS is trained with reinforcement learning, incorporating rewards that capture representativeness, diversity, query-adaptability and temporal coherence. We achieve state-of-the-art results on the MVS1K dataset, with inference time scaling linearly with the number of input video frames.
\end{abstract}



\maketitle

\begin{figure}[t!]
\centering
\vspace{1cm}
\includegraphics[width=\columnwidth]{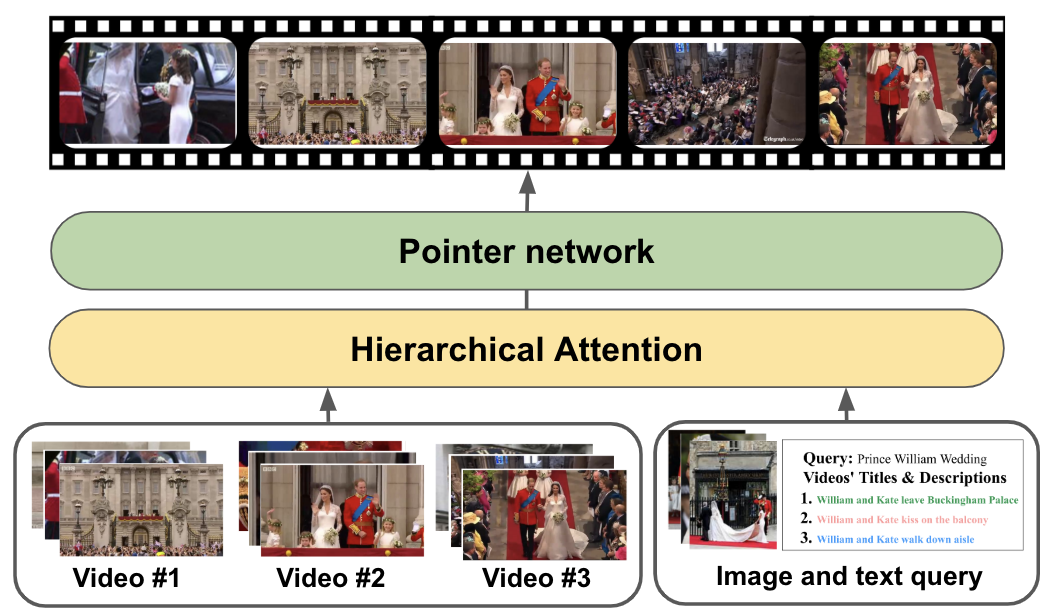}

\caption{Overview of the Deep Query-Aware Multi-Video Summarization (DeepQAMVS) model.}
\label{fig:teaser}
\vspace{0.2cm}
\end{figure}
 
\section{Introduction}\label{intro}
From Snapchat and Youtube to Twitter, Facebook and ByteDance, video sharing has influenced social media significantly over the past years. Video views increased over $99\%$ on YouTube and $258\%$ on Facebook, in just a single year\footnote{\url{https://www.wyzowl.com/video-social-media-2020/}}. To date, more than 5 billion videos have been shared on Youtube, where users daily spend 1 billion hours watching the uploaded content\footnote{\url{https://www.omnicoreagency.com/youtube-statistics/}}. Facebook also reached 100 million hours of video watching every day\footnote{\url{https://99firms.com/blog/facebook-video-statistics/}}. Given a query, current video search engines return hundreds of videos, often redundant and difficult for the user to comprehend without spending a significant amount of time and effort to find the information of interest. To effectively tackle this issue, \textit{Query-Aware Multi-Video Summarization (QAMVS)} methods select a subset of frames from the retrieved videos and form a concise topic-related summary conditioned on the user search intent \cite{Ji2017QueryAwareSC,Ji2018Archetypal}.

A compelling summary should be (1) concise, (2) representative of the query-relevant events, and (3) chronologically sound. Naively applying traditional single video summarization (SVS) techniques results in suboptimal summaries, as SVS methods fail to capture all aforementioned criteria. Overall, QAMVS is more challenging than SVS. First, QAMVS needs to ensure temporal coherence, a non-trivial task since the frames are selected from multiple different videos. In contrast, for SVS the chronological order is given by the video frame order. Secondly, QAMVS methods need to filter large noisy content as videos contain a lot of query-irrelevant information. Hence, QAMVS involves modeling the interactions between two or more modalities, \ie the set of videos and the query contents. In contrast, a clustering formulation optimizing for the summary diversity yields good results for SVS. 

Prior work relies on multi-stage pipelines to sequentially optimize for the aforementioned criteria. First, a set of candidate frames is selected following graph-based \cite{chu2015video,Ji2018Hypergraph,Kim2014JointSO}, decomposition-based \cite{Ji2017QueryAwareSC,Ji2018Archetypal,Panda2017DiversityAwareMS} or learning-based \cite{nie2015perceptual,wang2009multi} methods. Next, the list of frames is refined to be query-adaptive by ignoring frames that are dissimilar to a set of web-images retrieved with the same query \cite{Ji2017QueryAwareSC,Ji2018Hypergraph,Ji2018Archetypal,Kim2014JointSO}. Finally, the selected frames are ordered to form a coherent summary, either based on importance scores assigned at the video level \cite{Panda2017DiversityAwareMS,wang2009multi} or by topic-closeness \cite{Ji2017QueryAwareSC,Ji2018Hypergraph,Ji2018Archetypal}.
Due to the sequential nature of these methods we observe significant shortcomings: 
(1) multi-stage procedures result in error propagation; (2) existing methods have polynomial complexity with respect to the size of the video set and the video lengths, and (3) the use of multi-modal meta-data is often limited to candidate frame selection instead of guiding the summarization in every step.

To address these shortcomings, in this work, we propose a \textit{unified end-to-end trainable model} for the QAMVS task. Our architecture (summarized in Figure \ref{fig:teaser}) is a hierarchical attention-based sequence-to-sequence model which significantly reduces the computational complexity from polynomial to linear compared to the current state-of-the-art methods and alleviates error propagation due to being a unified approach. We achieve this via a pointer network, which selects the frames to include in the summary, thus removing the burden of rearranging the frames in a separate subsequent step. The attention of the pointer network factorizes over three distributions, each collecting evidence from a different modality, guiding the summarization process in every step. To address the challenge of limited ground truth supervision, we train our model using reinforcement learning, incorporating representativeness, diversity, query-adaptability and temporal coherence rewards.

The key contributions of this work are summarized as follows: (1) We design a novel end-to-end Query-Aware Multi-Video Summarization (DeepQAMVS) framework that jointly optimizes multiple crucial criteria of this challenging task: (i) conciseness, (ii) chronological soundness and (iii) representativeness of all query-related events. (2) We adopt pointer networks to remove the burden of rearranging the selected frames towards forming a chronologically coherent summary and design a hierarchical attention mechanism that models the cross-modal semantic dependencies between the videos and the query, achieving state-of-the-art performance. (3) We employ reinforcement learning to avoid over-fitting to the limited ground-truth data. We introduce two novel rewards that capture query-adaptability and temporal coherence.
We conduct extensive experiments on the challenging MVS1K dataset. Quantitative and qualitative analysis shows that our model achieves state-of-the-art results and generates visually coherent summaries.

\section{Related Work}
We cover related work on single video summarization (SVS), multi-video summarization (MVS) and pointer networks (PN). 
\subsection{Single Video Summarization}
Both \textit{supervised} and \textit{unsupervised} methods have been proposed for the  SVS task. On the \textit{supervised} side, methods involve category-specific classifiers for importance scoring of different video segments \cite{potapov2014category,sun2014ranking}, sequential determinantal point processes \cite{Gong2014DiverseSS,sharghi2018improving,sharghi2016query}, LSTMs \cite{Liu2019ICIP,Sahrawat2019,zhang2016video}, encoder-decoder architectures \cite{Cai2018ECCV,encode-decode2017}, memory networks \cite{Feng2018ExtractiveVS} and semantic aware techniques which include video descriptors \cite{Wei2018VideoSV}, vision-language embeddings \cite{plummer2017enhancing,Vasudevan2017QueryadaptiveVS} and text-summarization metrics \cite{yeung2014videoset}.
Instead, \textit{unsupervised} methods rely on low-level visual features to determine the important parts of a video. Strategies include clustering \cite{de2011vsumm,guan2014top,Pan2019}, maximal bi-clique finding \cite{chu2015video}, energy minimization \cite{pritch2007webcam} and sparse-coding \cite{cong2011towards,dornaika2015decremental,elhamifar2012see,zhao2014quasi}. Recently, convolutional models \cite{Rochan2018ECCV}, generative adversarial networks \cite{Fu2019WACV,mahasseni2017unsupervised,Rochan2019CVPR,Zhang2018QueryConditionedTA,Zhang2018DTRGANDT} and reinforcement learning \cite{Lan2018FFNetVF,Paulus2018ADR,zhang2019,zhou2018deep} have shown compelling results on the SVS task. 

Using queries to guide the summary has been explored in SVS. Proposed methods condition the summary generation on the textual query embedding \cite{zhang2019,Zhang2018QueryConditionedTA}, learn common textual-visual co-embeddings for both the query and the frames \cite{Vasudevan2017QueryadaptiveVS}, or enrich the visual features with textual ones obtained from dense textual shot annotations \cite{sharghi2016query,sharghi2017query}. As current multi-video datasets contain video-level titles/descriptions and abstract queries (\eg retirement, wedding, terror attack), the aforementioned methods are not applicable. Instead, we use the query to retrieve a set of web-images that represent its major sub-concepts/sub-events and use these images to condition the summary generation process. Note that query-adaptability is more critical in the case of MVS due to large irrelevant content across different videos.

In general, SVS methods that operate on a single long video obtained by concatenating all videos to be summarized, such as $k$-means \cite{de2011vsumm} and Dominant Set Clustering (DSC) \cite{besiris2009combining}, also result in lower performance than methods designed specifically for the QAMVS problem. These methods first form clusters of frames, select centroids as candidate frames and then compute diversity to eliminate similar keyframes before generating the final summary. Due to the lack of an ordering mechanism, SVS methods result in low consistency across selected frames that reduces readability and smoothness of the overall summary, affecting significantly the user viewing experience \cite{de2020content}. Nevertheless, to emphasize the importance of designing techniques that tackle QAMVS specifically, we also report results for SVS approaches in our evaluation. 

\begin{figure*}[t!]
    \centering
    \includegraphics[width=0.95\linewidth]{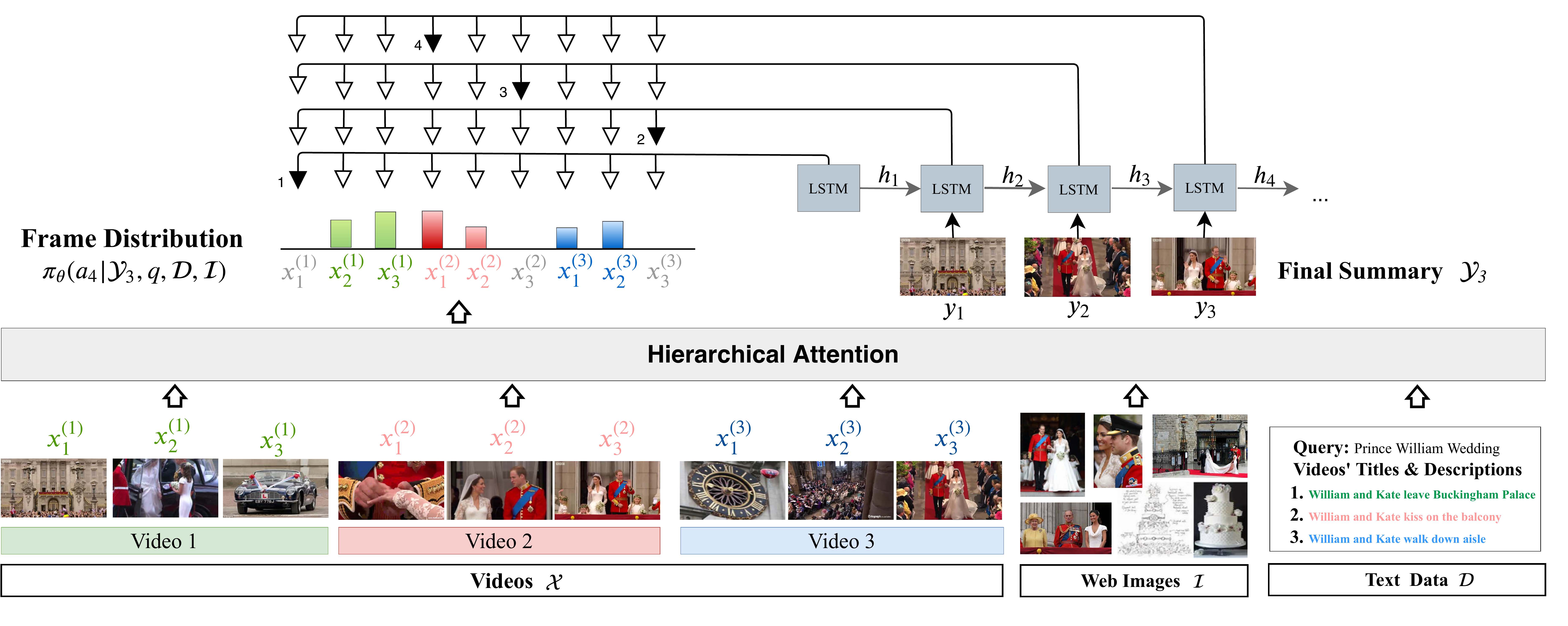}
        \caption{Overview of the policy network. DeepQAMVS is modeled as a Pointer Network with Hierarchical Attention (Figure \ref{fig:hierarchical_attention}). The policy $\pi_{\theta}(a_{t}|\mathcal{Y}_{t-1}, q, \mathcal{D}, \mathcal{I})$ is constructed by gathering evidence from the videos, the query images and the textual data. During inference, the frame  with the highest probability from the video collection is copied into the final summary $\mathcal{Y}_{L}$.}
    \label{fig:inference}
\end{figure*}
\subsection{Multi-Video Summarization} 
Applications range from multi-view summarization aiming at summarizing videos captured for the same scene with several dynamically moving cameras (\eg in surveillance) \cite{hoshen2014wisdom,mahapatra2016mvs,ou2014line,ou2014communication,zhu2015multicamera}, and summarizing of  user-devices' videos \cite{arev2014automatic,nie2015perceptual,zhang2014probabilistic,zhang2014effective,zhang2013dynamic,zhang2012multi,zhang2016efficient} (\eg for cities hotspot preview \cite{zhang2012multi} or city navigation \cite{zhang2013dynamic}) to topic-related MVS (QAMVS) \cite{Ji2017QueryAwareSC,Ji2018Archetypal,Kim2014JointSO,nie2015perceptual,Panda2017DiversityAwareMS,wang2012event}. 
Early attempts to solve the QAMVS task applied techniques optimizing for diversity \cite{dale2012multi,hong2009event,kanehira2018aware,li2010multi1,li2010multi2,li2012video,li2016multimedia,li2011static,Panda2017DiversityAwareMS,Panda2017CollaborativeSO,wang2009multi,wang2012event}. However, methods that advocate for these metrics cannot solve the QAMVS task satisfactorily, as (1) unimportant yet diverse frames are selected due to the high amount of irrelevant information across the different videos and (2) frames are not ordered chronologically to make a coherent story. Nevertheless, to emphasize the importance of designing techniques that tackle QAMVS challenges specifically, we also report results for diversity-oriented approaches in our evaluation. More recent QAMVS methods can be divided into three categories: (1) graph-based, (2) decomposition-based, and (3) learning-based.

\textit{Graph-based methods} construct a graph of relationships between frames of different videos, from which the most representative ones are selected. For example, Kim \etal \cite{Kim2014JointSO} summarized query related videos by performing diversity ranking on top of the similarity graphs between query web-images and video frames, in order to reconstruct a storyline graph of query-relevant events. Ji \etal \cite{Ji2018Hypergraph} proposed a clustering-based procedure using a hyper-graph dominant set, followed by a refinement step to filter frames that are most dissimilar to the query web-images, and a final step where the remaining candidates are ordered based on topic closeness. 

\textit{Decomposition-based} approaches subsume weighted archetypal analysis and sparse-coding.  Ji \etal \cite{Ji2018Archetypal} proposed a two-stage approach, where the frames are first extracted using multimodal Weighted Archetypal Analysis (MWAA). Here, the weights are obtained from a graph fusing information from video frames, textual meta-data and query-dependent web-images. Next, the frames are chronologically ordered based on upload time and topic-closeness. 
Panda \etal \cite{Panda2017DiversityAwareMS} formulated QAMVS as a sparse coding program regularized with \textit{interestingness} and \textit{diversity} metrics, followed by ordering the frames using a video-relevance score.
While Panda \etal \cite{Panda2017DiversityAwareMS} did not account for query-adaptability, Ji \etal \cite{Ji2017QueryAwareSC} extended the latter with an additional regularization term enforcing the selected frames to be similar to the query web-images. To form the final summary, frames are then ordered chronologically by grouping them into events based on textual and visual similarity.

For \textit{learning-based} methods, Wang \etal \cite{wang2012event} proposed a multiple-instance learning approach to localize the tags into video shots and select the query-aware frames in accordance with the tags. Nie \etal \cite{nie2015perceptual} selected frames from semantically important regions and then use a probabilistic model to jointly optimize for multiple attributes such as aesthetics, coherence, and stability.

In contrast to previous approaches that propose modularized solutions, we design a unified end-to-end model for QAMVS to generate visually coherent summaries in an end-to-end fashion. 
\begin{figure*}[t!]
    \centering
    \includegraphics[width=0.95\linewidth]{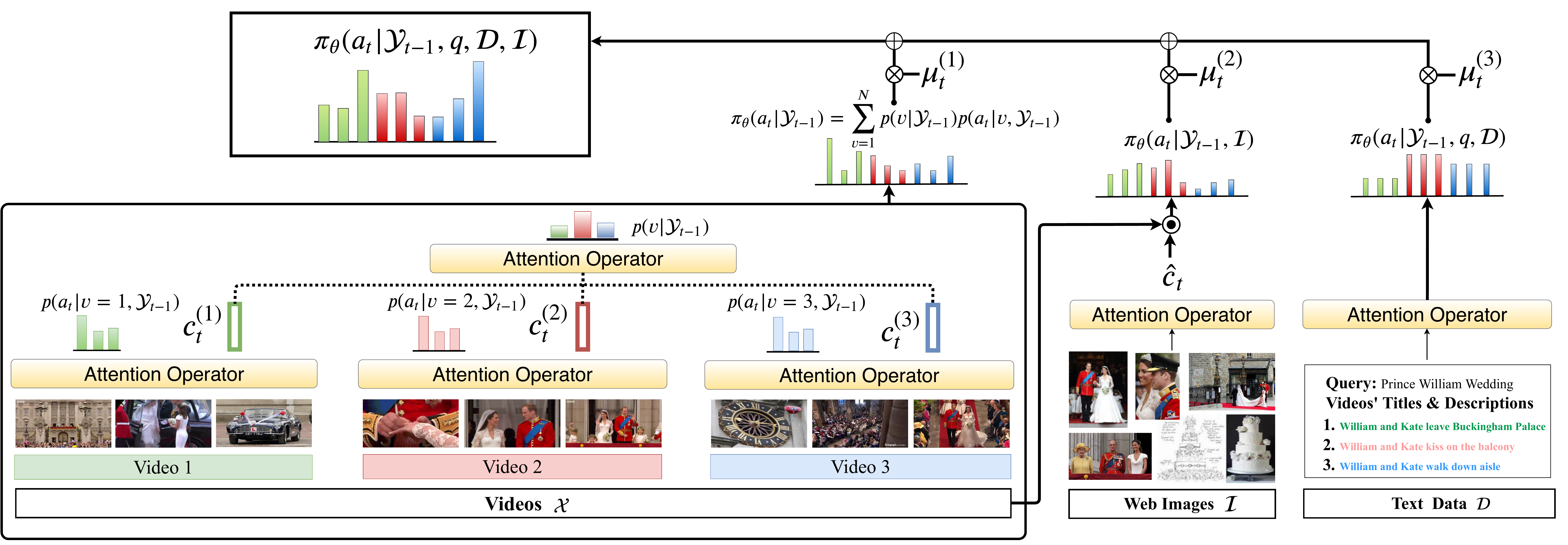}
    \caption{Illustration of DeepQAMVS's Hierarchical Attention $\pi_{\theta}(a_{t} \vert \mathcal{Y}_{t-1},q,\mathcal{D}, \mathcal{I})$.}
    \label{fig:hierarchical_attention}
\end{figure*}

\subsection{Pointer Networks}
Pointer Networks (PNs) have been applied to solve combinatorial optimization problems, \eg traveling-salesman \cite{bello2016neural} and language modeling tasks \cite{vinyals2015order}. 
At every time step, the output is constructed by iteratively copying an input item that is chosen by the pointer. This property is uniquely convenient for the QAMVS task. Our model is the first to use a Pointer Network for QAMVS. PNs, unlike other Seq2Seq models (\eg LSTM \cite{zhang2016video} or seqDPP \cite{Gong2014DiverseSS}), enable attending to any frame in any video at any time point. Hence, they naturally generate an ordered sequence of frames, while the attention mechanism fuses the multi-modal information to select the next best frame satisfying diversity, query-relevance and visual coherence (Figure \ref{fig:hierarchical_attention}). We train the Pointer Network in our model using reinforcement learning, as it is useful for tasks with limited labeled data \cite{li2016deep,he2016dual,bunel2018leveraging,liang2018memory,BoyanJMLR2000,sharaf2017structured,messaoud2020can}, as in the case of QAMVS.

\section{Proposed DeepQAMVS Model}
Given a collection of videos and images retrieved by searching with a common text query that encodes user preferences, the goal is to generate a topic-related summary for the videos. DeepQAMVS utilizes both web-images and textual meta-data. Web-images are particularly useful as they guide the summarization towards discarding irrelevant information (\textbf{image attention}). However, they might contain irrelevant information. To robustly ensure query-relevance, DeepQAMVS leverages multi-modal attention which enables the web-images and textual meta-data (\textbf{query attention}) to act as complementary information that guides the summarization process. In the following, we first formally define the problem, then introduce the proposed DeepQAMVS model.

\subsection{Problem Formulation}
Let $q$ be the semantic embedding of the textual query and let $\mathcal{I} = \{I_{1} , \cdots, I_{|\mathcal{I}|}\}$ refer to the set of web-image embeddings. We denote by
$\mathcal{X}^{(v)}=\{ x_{1}^{(v)}, \cdots, x_{|\mathcal{X}^{(v)}|}^{(v)}\}$
the set of frame embeddings from video $v \in \{1,\cdots,N\}$. Let $\mathcal{D}=\{ d^{(1)}, \cdots, d^{(N)}\}$ be the text embeddings of the videos' textual data, constructed by averaging the embeddings of the title and description for every video. The goal is to generate a summary $\mathcal{Y}_{L}=\{ y_{1}, \cdots,  y_{L} \}$ of $L$ frames selected from the input video frames, \ie $\mathcal{Y}_{L} \subset \mathcal{X} = \bigcup_{v} \mathcal{X}^{(v)}$.

Due to the sequential nature of the problem, \ie selecting the next candidate frame based on what has been selected so far, we formulate the QAMVS problem as a Markov Decision Process (MDP). Specifically, an agent operates in $t \in \{1, \cdots , L\}$ time-steps according to a policy $\pi_{\theta}(a_{t}|\mathcal{Y}_{t-1}, q,\mathcal{D},\mathcal{I})$ with trainable parameters $\theta$. The policy encodes the probability of selecting an action $a_{t}$ given the  state $\mathcal{Y}_{t-1}$, the query $q$, the text meta-data $\mathcal{D}$ and the web-images $\mathcal{I}$. The  state $\mathcal{Y}_{t-1}$ denotes the set of  frames that are already selected in the summary up to time step $t$. Note that $\mathcal{Y}_{0}= \emptyset$. The set of possible actions is the set of input frames after eliminating the ones that have already been selected in the summary, \ie $ a_{t} \in \mathcal{A}_{t}= \mathcal{X} \setminus \mathcal{Y}_{t-1}$.  We denote by $\mathcal{A}_{t}^{(v)} $ the set of valid actions corresponding to frames from video $v$.

We model the policy function $\pi_{\theta}(a_{t}| \mathcal{Y}_{t-1}, q,\mathcal{D},\mathcal{I} )$  as a pointer network with hierarchical attention, as illustrated in Figure \ref{fig:inference}. {\bf At inference} step $t$, the inputs ($\mathcal{X}$, $\mathcal{D}$ and $\mathcal{I}$), together with the state $\mathcal{Y}_{t-1}$, are used to compute the distribution $\pi_{\theta}(a_{t}|\mathcal{Y}_{t-1},q,\mathcal{D},\mathcal{I})$ over possible actions $a_{t}$, \ie over possible frames. The frame with the highest probability is then copied to the summary $\mathcal{Y}_t$. The process continues until a  summary of length $L$ is reached. 
Next, we describe the policy  $\pi_{\theta}(a_{t}|\mathcal{Y}_{t-1}, q,\mathcal{D},\mathcal{I})$. 
\subsection{DeepQAMVS Policy Network}\label{sec:PolicyNetwork}
Our proposed policy function models the cross-modal semantic dependencies between the videos, the text query and the web-images. More specifically, the policy network $\pi_{\theta}(a_{t}|\mathcal{Y}_{t-1},q,\mathcal{D},\mathcal{I})$ is the weighted combination of three distributions, video frame attention $\pi_{\theta}(a_{t}|\mathcal{Y}_{t-1})$, image attention $\pi_{\theta}(a_{t}|\mathcal{Y}_{t-1},\mathcal{I})$, and query attention $\pi_{\theta}(a_{t}|\mathcal{Y}_{t-1},q,\mathcal{D})$. 
Formally, 
\begin{equation} \label{eq:total_prob}
\begin{split}
\pi_{\theta}(a_{t}|\mathcal{Y}_{t-1},q,\mathcal{D},\mathcal{I}) =
\mu_{t}^{(1)} \pi_{\theta}(a_{t}|\mathcal{Y}_{t-1}) + \\
\mu_{t}^{(2)} \pi_{\theta}(a_{t}|\mathcal{Y}_{t-1}, \mathcal{I}) + 
\mu_{t}^{(3)} \pi_{\theta}(a_{t}|\mathcal{Y}_{t-1}, q, \mathcal{D}), 
\end{split}
\end{equation}
where $\mu_{t}^{(1)}$, $\mu_{t}^{(2)}$ and $\mu_{t}^{(3)}$ are learnable interpolation terms satisfying $ \mu_{t}^{(1)} +  \mu_{t}^{(2)} +  \mu_{t}^{(3)} = 1$. An illustration of the hierarchical attention is provided in Figure \ref{fig:hierarchical_attention}. In the following, we introduce each of these three distributions.\\

\noindent The \textbf{video frame attention} $\pi_{\theta}(a_{t}|\mathcal{Y}_{t-1})$ is modeled as a two-level attention, 
\ie at each time step $t$, video attention selects video $v$ and then selects a frame $a_{t}$ from video $v$:
\begin{align}\pi_{\theta}(a_{t}|\mathcal{Y}_{t-1})= \sum\limits_{v=1}^N p(v|\mathcal{Y}_{t-1})  p(a_{t}|v, \mathcal{Y}_{t-1}),\end{align} 
where $p(a_{t}|v,\mathcal{Y}_{t-1})$ is the probability of selecting a frame $a_{t}$ from video $v$, and $p(v|\mathcal{Y}_{t-1})$ is the distribution over the collection of videos.
We compute both probabilities via
\begin{align} 
&p(a_{t}|v, \mathcal{Y}_{t-1}),  c_{t}^{(v)} =  \text{Attention}( \mathcal{A}_{t}^{(v)} , \mathcal{Y}_{t-1}  ), \label{eq:frame_Attention} \\ &p(v|\mathcal{Y}_{t-1}),  c_{t}  =  \text{Attention}\left( \{ c_{t}^{ (1) }, \cdots , c_{t}^{ (N)} \} , \mathcal{Y}_{t-1}  \right). \label{eq:video_Attention}
\end{align}

The $\textit{Attention}$ operator, as well as the context vectors $c_{t}$ and $\{c_{t}^{ (v) }\}_{v=1}^{N}$ are defined below.
Intuitively, the two-level attention enables scaling to a large number of videos and video lengths since it decomposes a joint distribution into the product of two conditional distributions.\\

\noindent The \textbf{image attention} $\pi_{\theta}(a_{t}|\mathcal{Y}_{t-1},\mathcal{I})$ reflects the correlation between video frames and web-images. We first generate a context vector $\hat{c}_{t}$ encoding the most relevant information in the web-images at time $t$ given the current summary $\mathcal{Y}_{t-1}$:
\begin{equation} 
p(I|\mathcal{Y}_{t-1} ),\hat{c}_{t} =   \text{Attention}( \mathcal{I}, \mathcal{Y}_{t-1} ). 
\label{Frame_Dec_Attention}
\end{equation}
$\pi_{\theta}(a_{t}| \mathcal{Y}_{t-1}, \mathcal{I})$ is then obtained by transforming the dot product between $\hat{c}_{t}$ and the action representations, \ie representations from not previously selected frames,  into a distribution via a softmax.\\

\noindent The \textbf{query attention} 
$\pi_{\theta}(a_{t}|\mathcal{Y}_{t-1}, q, \mathcal{D})$ captures the correlation between the query $q$, the text data $\mathcal{D}$ and the summary at time $t$. For this, we first weigh every video's text embedding by its similarity to the query. Next, we compute an attention over the weighted embeddings, given the current summary $\mathcal{Y}_{t-1}$, via
\begin{align} 
\pi(a_{t}|\mathcal{Y}_{t-1}, q, \mathcal{D}), \tilde{c_{t}} \!=\! \text{Attention}\left( (q^{T} d^{(v)}) d^{(v)} , \mathcal{Y}_{t-1} \right). \label{eq:text_dist}
\end{align}

\noindent The interpolation weights in Eq.~\eqref{eq:total_prob} can be obtained by attending over the modalities' context vectors, $c_{t}$, $\hat{c}_{t}$ and $\tilde{c_{t}}$:
\begin{equation}
[\mu_{t}^{(1)},\mu_{t}^{(2)}, \mu_{t}^{(3)}], \cdot =  \text{Attention}( \{c_{t},\hat{c}_{t},\mathrm{MLP}(\tilde{c_{t}})\}, \mathcal{Y}_{t-1}),
\end{equation} 
where $\mathrm{MLP}$ is a multi-layer perceptron used to unify the dimensions of the three context vectors. We observe that if $c_{t}$ and $\hat{c}_{t}$ are similar, their weights $\mu_{t}^{(1)}$ and $\mu_{t}^{(2)}$ are close, else more weight is given to \textit{video attention}. \\

\begin{figure}[t!]
\includegraphics[width=0.75\columnwidth]{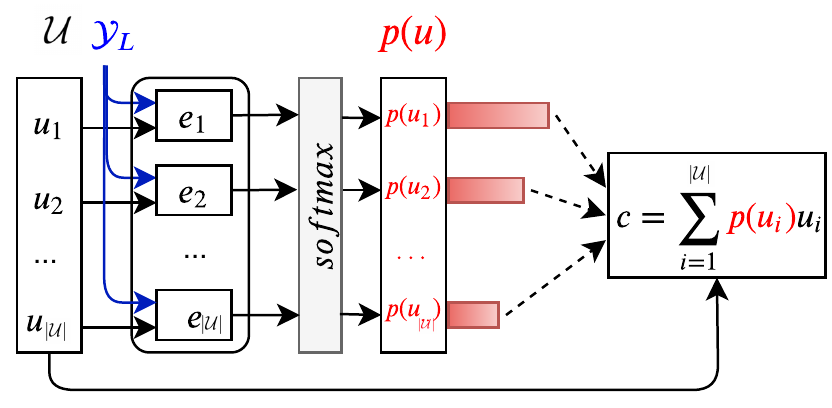}
\caption{The \textit{Attention} operator.}
\label{fig:attention}
\end{figure}
\noindent The \textbf{$\text{Attention}$ operator}, illustrated in Figure \ref{fig:attention} and used multiple times above, takes as input a sequence of vectors $\mathcal{U} = \{u_{i}\}_{i=1}^{\vert \mathcal{U} \vert}$ with $u_{i} \in \mathbb{R}^{m}$ and the summary $\mathcal{Y}_{t-1}$, embedded by an LSTM into a hidden state $h_{t} \in \mathbb{R}^{n}$. The Attention operator provides as output a distribution $p(u_{i})$ over the vectors $\{u_{i}\}_{i=1}^{\vert \mathcal{U} \vert}$ and a context vector $c$ as a linear combination of elements in $\mathcal{U}$ by conditioning them on $h_{t}$:
\begin{gather} 
e_{i}= w_{1}^{T} \text{tanh}(W_{2}[u_{i} ; h_{t} ]), \nonumber 
\\
p(u_{1}),\cdots,p(u_{\vert \mathcal{U} \vert}) = \text{Softmax}([e_1,\cdots,e_{\vert \mathcal{U} \vert}]),\label{eq:att_2}\\
c = \sum_{i=1}^{\vert \mathcal{U} \vert} p(u_{i}) u_{i},\label{eq:att_3}
\end{gather}
where $w_1 \in \mathbb{R}^{n}$ and $W_2 \in \mathbb{R}^{n \times (n+m)}$ are trainable weight parameters. The outputs of the Attention operator are the probabilities given in Eq.~\eqref{eq:att_2} and the context vector $c$ given in Eq.~\eqref{eq:att_3}. \\

\noindent \textbf{Embeddings:} The video frames $\mathcal{X}$ are embedded with a pre-trained CNN followed by a BiLSTM network. Web-images $\mathcal{I}$ are encoded with the same CNN. Textual embeddings $\mathcal{D}$ are computed for every video by averaging Glove word embeddings \cite{pennington2014glove} from its associated title and description. Note that we normalize all embeddings.

\subsection{Training with Policy Gradient}
Due to the limited annotated data and the subjectivity of the ground truth summaries, we train our model via reinforcement learning. The goal is to learn the policy $\pi_{\theta}(a_{t}|\mathcal{Y}_{t-1},q,\mathcal{D},\mathcal{I})$ by maximizing the expected reward $J(\theta) = \mathbb{E}_{\pi_{\theta}}[R(\mathcal{Y}_{L})]$ during training, where $R(\mathcal{Y}_{L})$ denotes the reward function computed for a summary $\mathcal{Y}_{L}$. Following REINFORCE \cite{williams1992simple}, we approximate the expectation by running the agent for $M$ episodes for a batch of videos and then taking the average gradient. To reduce variance, we use a moving average of the rewards as a computationally efficient baseline. 

The reward $R = \beta_{1} R_\text{div}+ \beta_{2} R_\text{rep}+ \beta_{3} R_\text{query} + \beta_{4} R_\text{coh}
$ is composed of four terms, measuring the diversity ($R_\text{div}$), representativeness ($R_\text{rep}$), query-adaptability ($R_\text{query}$) and temporal coherence ($R_\text{coh}$).
Hyperparameters $\{\beta_i\}_{i=1}^4$ are weights associated to different rewards.
Note that we use the same diversity and representativeness rewards as \citet{zhou2018deep}. In addition, we introduce two novel rewards, query-adaptability and temporal coherence, to accommodate  the QAMVS task. To keep the rewards in the same range, we use (1) \textit{dot product} as a similarity metric in $R_\text{coh}$ to balance out $R_\text{div}$ and (2) a similar form to $R_\text{rep}$ for $R_\text{query}$.

\noindent The \textbf{Diversity Reward} measures the dissimilarity among the selected frames in the feature space via
\begin{equation}
R_\text{div}(\mathcal{Y}_{L})=\frac{1}{L(L-1)}  \sum_{  \substack{y_{t},y_{t'} \in \mathcal{Y}_{L} \\ t \neq t'}  } \left(1-y_{t}^{T} y_{t^\prime}\right).
\end{equation}
Intuitively, the more dissimilar the selected frames to each other, the higher the diversity reward the agent receives.

\noindent The \textbf{Representativeness Reward} measures how well the generated summary represents the main events occurring in the collection of videos. Thus, the reward is higher when the selected frames are closer to the cluster centers. Formally, 
\begin{equation}
R_\text{rep}(\mathcal{Y}_{L})\!=\!\exp(- \frac{1}{| \mathcal{X}| } \sum\limits_{x \in \mathcal{X} }\min\limits_{y_{t} \in \mathcal{Y}_{L} } \|x-y_{t} \|^{2}).
\end{equation}

\noindent The \textbf{Query-Adaptability Reward}\footnote{Other explored forms include $ R_\text{query}(\mathcal{Y}_{L})=-\frac{1}{|\mathcal{I}|}\!\sum\limits_{I \in \mathcal{I}} \min\limits_{y_{t} \in \mathcal{Y}_{L}}  \| y_{t} - I \|^{2}$ and $R_\text{query}(\mathcal{Y}_{L})=-\frac{1}{L}\!\sum\limits_{y_t \in \mathcal{Y}_{L}}\| y_{t} - \frac{1}{|\mathcal{I}|}\sum\limits_{I\in \mathcal{I}} I \|^{2}$. We found the formulation in Eq. \eqref{eq:query_reward} to work best.} encourages the model to select the summary frames to be similar to the web-images $\mathcal{I}$ via \begin{equation} \label{eq:query_reward}
R_\text{query}(\mathcal{Y}_{L}) = \exp \left( - \frac{1}{L} \sum_{y_t \in \mathcal{Y}_{L}} \min_{I \in \mathcal{I}} \Vert y_{t}-I  \Vert^{2}  \right).
\end{equation}

\noindent The \textbf{Temporal Coherence Reward} encourages the visual coherence of the generated summary via 
\begin{equation}
R_\text{coh}(\mathcal{Y}_{L}) = \frac{1}{L} \sum_{y_t \in \mathcal{Y}_{L}} \rho( y_{t} ),
\label{eq:temporal_coherence}
\end{equation}
where $\rho(y_t)$ is calculated by adding up the correlation between two consecutive frames:
\begin{align}
\rho(y_t) = \frac{1}{2} \sum_{k\in\{\pm 1\}} y^T_{t} y_{t+k}.\label{eq:corr}
\end{align}
Hence, the more correlated the neighboring frames, the higher the temporal coherence reward. Note that optimizing for visual/temporal coherence, \ie smoothness of the transitions, is just a proxy for chronological soundness, which is a much harder problem.

\section{Experimental Setup}
We describe experimental details, such as the evaluation dataset and metrics, and present quantitative and qualitative results, comparing the proposed DeepQAMVS model with several baselines.
Our experiments aim to show that (1) SVS methods cannot properly address QAMVS and multi-stage MVS procedures result in lower performance than a unified system (sections \ref{sec:exper} and \ref{sec:qualitative}), (2) the use of multi-modal information is crucial in guiding the summarization process (section \ref{sec:ablation}), (3) the introduced novel temporal coherence reward generates more visually coherent summaries (section \ref{sec:coherence}) and (4) our method reduces the computational complexity compared to the current state-of-the-art methods (section \ref{sec:runtime}).

\subsection{Experimental Settings}
\noindent \textbf{Dataset:} We perform our experiments on the MVS1K dataset \cite{Ji2017QueryAwareSC}.
MVS1K is a collection of $1000$ videos on $10$ queries (events), with associated web-images, video titles, and their text descriptions. Each query has $4$ different user summaries, serving as a ground truth. Table~\ref{tab:dataset_description} lists the events, the query used to retrieve them, the number of videos and query web-images for each event as well as the total number of input frames across all videos. Each video is associated with a title and a text description.
We use the features introduced by Ji \etal \cite{Ji2017QueryAwareSC}. The dimensionality of the video frame and web-image embeddings is $4352$. The embeddings are composed of a $4096$ dimensional VGGNet-19~\cite{simonyan2014very} (trained  on ImageNet) CNN feature vector concatenated with a $256$ dimensional HSV color histogram feature vector. These embeddings are reduced to a vector of length 256 through a fully connected layer. The input frames to the model are selected such that they represent the segment centers obtained using the shot boundary detection algorithm~\cite{yuan2007formal}. 
The textual features (titles and descriptions) and the query are Glove embeddings \cite{pennington2014glove} of dimension $100$. We set the hidden state dimension of the LSTM and the pointer network to be $256$ and $32$, respectively.\\

\noindent \textbf{Evaluation Metrics:} To compare with previous work, generated summaries are assessed using F1-score, averaged over the ground truth user summaries. Following prior work~\cite{Ji2017QueryAwareSC,Ji2018Hypergraph,Ji2018Archetypal}, two frames are considered to match when the pixel-level euclidean distance is smaller than a predefined threshold of 0.6. \\

\noindent\textbf{Training  Details:} We train using a 10-fold cross-validation scheme. Specifically, for evaluating each event, we use the remaining 9 events as training data. During training, we use a batch size of $32$, where each sample consists of $10$ randomly sampled videos per event. We limit the number of video combinations to $4000$ for every event. This large number of random combinations allowed us to avoid overfitting despite the small number of events\footnote{Besides the proposed reinforcement learning framework, we experimented with training in a supervised fashion, however, we could not avoid over-fitting.}. We optimize with Adam, $0.01$ learning rate and $\ell_2$ regularization.
During testing, we use all the videos associated with an event in the test set. 
Since the diversity $R_\text{div}$ and representativeness $R_\text{rep}$ reward on one side, and the coherence reward $R_\text{coh}$ on the other side are contradictory, \ie $R_\text{div}$ and $R_\text{rep}$ encourage the selection of diverse frames while $R_\text{coh}$ is high when the summary is smooth as measured by the similarity of the neighboring frames, we use a training schedule: (1) We set $\beta_{1}=\beta_{2}=\beta_{3}=1/3$ and $\beta_{4}=0$ for $60$ epochs. (2) Then, set $\beta_{1}=\beta_{2}=\beta_{3}=\beta_{4}=1/4$ for $30$ additional epochs. We also experiment with different summary lengths $L \in \{30,50,60\}$.

\begin{table}[t!]
\centering
\caption{Dataset Characteristics}
\vspace{-0.2cm}
\resizebox{\columnwidth}{!}{
\begin{tabular}{ccccc} 
\toprule
\textbf{{Query ID}} & \textbf{{Query}}  & \textbf{{\# Videos}} & \textbf{{\# Frames}} & \textbf{{\# Images}}  \\ \midrule
{1} & {Britains Prince William wedding 2011}   & {90} & {1124}    & {324}   \\
{2} & {Prince death 2016} & {104} & {1549}  &  {142}  \\
{3} & {NASA discovers Earth-like planet}  & {100} & {1349} & {226} \\
{4} & {American government shut-down 2013}  &  {82} & {962}  &  {177}  \\
{5} & {Malaysia Airline MH370}  & {109} & {1330}  & { 435}  \\
{6} & {FIFA corruption scandal 2015}  &  {90} & {785} & {177}  \\
{7} & {Obama re-election 2012}  &  {85} & {1263} & {207}  \\
{8} & {Alpha go vs Lee Sedo}  &  {84} & {976} & {118}  \\
{9} & {Kobe Bryant retirement}  & {109}     & {1140} & {221}  \\
{10} & {Paris terror attacks}  &  {83}  & {857}  & {651}  \\ \midrule 
\textbf{{Total}} &  -  &  {936} & - & {2678}   \\
\bottomrule
\end{tabular}
}
\label{tab:dataset_description}
\end{table}
\begin{table*}[t!]
\centering
\caption{Comparison of our approach against baselines (F1 score).}
\vspace{-0.1cm}
\resizebox{\textwidth}{!}{%
\begin{tabular}{c|cccccccccccc} 
\toprule
& \textbf{Query ID} &\textbf{ 1} & \textbf{2} & \textbf{3} & \textbf{4} & \textbf{5} & \textbf{6}& \textbf{7} & \textbf{8} & \textbf{9} & \textbf{10} & \textbf{AVG}  \\
\midrule
\multirow{5}{*}{\rotatebox[origin=c]{90}{\textbf{SVS}}}&\textbf{ k-means}     & .576  & .552 & \textbf{.568} & .336 & .457 & .525 & .651 & .278 & .384 & .337 & .466 \\
& \textbf{DSC} & .578  & .472 & .399 & .530 & .407 & .494 & .533 & .485 & .529 & .471 & .490  \\ 
 & \textbf{MSR}   & .472  & .391 & .370 & .414 & .396 & .355 & .418 & .234 & .384 & .288 & .372  \\
 & \textbf{SUM-GAN}   & .620$\pm$.035  &  .481$\pm$.028 & .519$\pm$.034  & .501$\pm$.038  & .413$\pm$.022  & .455$\pm$.048 & .458$\pm$.059  & .459$\pm$.041 & .510$\pm$.021 & .395$\pm$.056  & .486$\pm$ .075 \\ 
  & \textbf{DSN}   & .529$\pm$.019  & .327$\pm$.062  &  .478$\pm$.036  & .407$\pm$.026  & .325$\pm$.042  & .453$\pm$.033 & .616$\pm$.028  & .375$\pm$.022  & .469$\pm$.021 & .384$\pm$.016 & .436$\pm$.093  \\ 
 \midrule
\multirow{9}{*}{\rotatebox[origin=c]{90}{\textbf{MVS}}}&\textbf{QUASC}& .520  & .513 & .400 & \textbf{.570} & .513 & \textbf{.538} & .623 & .439 & \textbf{.709} & \textbf{.588} & .544  \\
&  \textbf{MVS-HDS}  & .660  & .552 & .475 & .526 & .495 & .520 & .642 & .469 & .633 & .581 & .555 \\
& \textbf{MWAA}  & .705  & \textbf{.610} & .553 & .511 & \textbf{.563} & .466 & \textbf{.664} & .483 & .611 & .379 & .555 \\ 
& \textbf{Random-50} & .600$\pm$.070 & .349$\pm$.088 & .288$\pm$.047 & .492$\pm$.131 & .255$\pm$.074 & .352$\pm$.096 & .265$\pm$.099 &  .429$\pm$.109& .326$\pm$.109 & .284$\pm$.064  & .364$\pm$.089\\
& \textbf{Ours-30} & $.570\!\pm\!.013$  & .491$\pm$.037 & .421$\pm$.084 & .519$\pm$.017 & .458$\pm$.054 & .476$\pm$.030 & .369$\pm$.036 & .372$\pm$.014 & .403$\pm$.017 & .368$\pm$.041 & .446$\pm$.022 \\  
& \textbf{Ours-50} & .706$\pm$.018  & .563$\pm$.035 & .525$\pm$.017 & .553$\pm$.026 & .549$\pm$.014 & .486$\pm$.032 & .524$\pm$.015 & \textbf{.486$\pm$.022}  & .690$\pm$.015 & .542$\pm$.022 & .561$\pm$.005 \\
& \textbf{Ours-60}&\textbf{ .722$\pm$.019}  & .530$\pm$.046 & .495$\pm$.009& .508$\pm$.015 & .541$\pm$.036 & .487$\pm$.014& .614$\pm$.026 & .474$\pm$.015 & .674$\pm$.025& .573$\pm$.019 & .562$\pm$.004\\ 
& \textbf{Ours-best}  & \textbf{ .722$\pm$.019} & .563$\pm$.035 & .525$\pm$.017 & .553$\pm$.026 & .549$\pm$.014 & .487$\pm$.014 & .614$\pm$.026 & \textbf{.486$\pm$.022} & .690$\pm$.015 & .573$\pm$.019 & \textbf{.576$\pm$.017} \\
\bottomrule \end{tabular}
}
\label{tab:experiments}
\end{table*}

\begin{table}[t]
\centering
\caption{Summary length (\# frames) across methods.}
\label{wrap-tab:1}
\vspace{-0.1cm}
\resizebox{\columnwidth}{!}{%
\begin{tabular}{c|cccccccccccc}
\toprule
&\textbf{Query ID} &\textbf{ 1} & \textbf{2} & \textbf{3} & \textbf{4} & \textbf{5} & \textbf{6}& \textbf{7} & \textbf{8} & \textbf{9} & \textbf{10} & \textbf{AVG}  \\
\midrule
\multirow{5}{*}{\rotatebox[origin=c]{90}{\textbf{ SVS}}}&\textbf{k-means} & 48  & 51 & 59 & 51 & 63 & 47 & 48 & 36 & 39 & 28 & 47.0 \\
&\textbf{DSC}  & 42  & 47 & 34 & 39 & 52 & 46 & 55 & 41 & 41 & 41 & 43.8   \\
&\textbf{MSR}    & 48  & 51 & 59 & 51 & 63 & 47 & 48 & 36 & 39 & 28 & 47.0  \\ 
&\textbf{SUM-GAN}    & 60  & 60 & 60 & 60 & 60 & 60 &60  & 60 & 60 & 60 &  60.0 \\ 
&\textbf{DSN}    & 60  & 60 & 60 &  60&  60&  60& 60 & 60  & 60 & 60 & 60.0  \\ 
\midrule
\multirow{3}{*}{\rotatebox[origin=c]{90}{ \textbf{MVS}}}&\textbf{QUASC}  & 33  & 57 & 21 & 55 & 48 & 41 & 59 & 52 & 51 & 56 & 47.3  \\
&\textbf{MVS-HDS} & 51   & 60 & 48 & 48 & 60 & 44 & 58 & 54 & 60 & 50 & 53.3\\ 
&\textbf{MWAA}& 49  & 75 & 46  & 39 & 49 & 37 & 60 & 36  & 39 & 39 & 46.9 \\ &\textbf{Ours-best} & 60  & 50 & 50  & 50 & 50 & 60& 60 & 50  & 50 & 60 & 53.0 \\ 
\bottomrule
\end{tabular}
}
\label{tab:baselines_frames}
\end{table}

\subsection{Experimental Results}\label{sec:exper}

We compare DeepQAMVS to five SVS baselines operating on the concatenated videos. We chose the SVS baselines such that they represent the main trends in unsupervised summarization: 
\begin{itemize}[leftmargin=*]
\item \textbf{$K$-means} \cite{de2011vsumm}: SVS method that clusters all video frames and then selects the  one closest to the cluster centers as summary frames ($k=9$).

\item \textbf{DSC} \cite{besiris2009combining}:  Dominant Set Clustering (DSC) is a graph-based clustering method where a dominant set algorithm is used to extract the summary frames.

\item \textbf{MSR} \cite{besiris2009combining}: Minimum Sparse Reconstruction (MSR) is a decomposition based approach, which formulates video summarization as a minimum sparse reconstruction.

\item \textbf{SUM-GAN} \cite{mahasseni2017unsupervised}: An adversarial LSTM model, where the generator is an autoencoder LSTM aiming at first selecting the summary frames then reconstructing the original video based on them, and the discriminator is trained to distinguish between the reconstructed video and the original one. 

\item \textbf{DSN} \cite{zhou2018deep}: uses a RNN trained with deep reinforcement learning with diversity and representativeness rewards. 
\end{itemize}

\noindent Moreover, we compare with four state-of-the-art QAMVS baselines: 
\begin{itemize}[leftmargin=*]
\item \textbf{QUASC} \cite{Ji2017QueryAwareSC}: QUASC is a sparse coding program regularized with interestingness, diversity and query-relevance metrics, followed by ordering the frames chronologically by grouping them into events based on textual and visual similarity.

\item \textbf{MWAA} \cite{Ji2018Archetypal}: MWAA is a two-stage approach, where the frames are first extracted using multi-modal Weighted Archetypal Analysis (MWAA), and then are chronologically ordered based on upload time and topic-closeness. 

\item \textbf{MVS-HDS} \cite{Ji2018Hypergraph}: MVS-HDS is a clustering-based procedure using a Hyper-graph Dominant Set, followed by a refinement step to filter frames that are most dissimilar to the query web-images, and a final step where the remaining candidates are ordered based on topic closeness. 

\item \textbf{Random-50}: We also compare our method against a randomly generated summary with length $50$.
\end{itemize}

We present quantitative results of our approach in Table~\ref{tab:experiments} and the number of summary frames selected by each approach in Table~\ref{tab:baselines_frames}. More specifically, in Table~\ref{tab:experiments}, the reported numbers represent the mean and standard deviation obtained from $5$ rounds of experiments. We report the F1-scores for summaries of length 30 (\textit{ours-30}), 50 (\textit{ours-50}) and 60 (\textit{ours-60}), as well as the best obtained score (\textit{ours-best}) when selecting the best summary length for every event. We observe that SVS methods have in general lower performance than MVS methods. In addition, our proposed end-to-end DeepQAMVS model, on average, outperforms all baselines. 

\subsection{Ablation Study}\label{sec:ablation}
We present an ablation study, examining the effect of different rewards and attention mechanisms in Table \ref{tab:ablation_study}. We evaluate the average F1-score across all the events for the following combinations of the attention modalities: (1) only video frame attention ($\mu_t^{(2)}=\mu_t^{(3)}=0$); (2) video frame and image attention ($ \mu_t^{(3)}=0$); (3) video frame and query attention ($\mu_t^{(2)}=0$); and (4) video frame, image and query attention ($\mu_t^{(i)} \neq 0$, $\forall \, i \in \{1,2,3\}$). We also investigate the effect of the different rewards by incrementally adding the reward terms including, (1) $R_\text{div}$ ($\beta_2=\beta_3=\beta_4=0$); (2) $R_\text{div}$ and $R_\text{query}$ ($\beta_3=\beta_4=0$); (3) $R_\text{div}$, $R_\text{query}$ and $R_\text{coh}$ ($\beta_4=0$); and (4) all the rewards, \ie  $\beta_i \neq 0$, $\forall i \in \{1,\cdots,4\}$. Note that we do not add the query reward $R_\text{query}$ when testing with attention terms that do not include the image attention ($-$ in Table~\ref{tab:ablation_study}).

When considering all forms of attention (last column), we found that $R_\text{div}$ has barely improved the F1-score. In contrast, including $R_\text{query}$,  helped improve the quality of the summary while adding the coherency reward $R_\text{coh}$ did not lead to a consistent increase of the F1-score. This is expected as the ground truth summary consists of an unordered set of frames. However, as demonstrated by the user study below, $R_\text{coh}$ helped generating more visually coherent summaries. Across the different combinations of rewards, we observe that the combination of video frame attention and image attention (column 3) yields overall a higher F1-score than the combination of video frame attention and query attention (column 4). This is due to video descriptions being noisy and associated with the whole video, unlike the web-images, which are embedded in the same space as the frames and hence better capture query-adaptability. The best results are obtained by using all the attention terms (last column), demonstrating the complementary properties of the multimodal information. 

\begin{table}[t!]
\centering
\caption{Ablation study on attention and rewards ($L=60$).}
\vspace{-0.1cm}
\setlength\tabcolsep{7pt}
\resizebox{\columnwidth}{!}{%
\begin{tabular}{c|cccc}
\toprule
\diagbox{Reward}{Attention} & 
\begin{tabular}[c]{@{}l@{}}$\mu_t^{(2)}=0$\\ $\mu_t^{(3)}=0$\end{tabular} &$\mu_t^{(3)}=0$&$\mu_t^{(2)}=0$&$\mu_t^{(i)}\neq 0$ \\
\midrule
$\beta_2=\beta_3=\beta_4=0$ & .323$\pm$ .013 & .559$\pm$.008& .374$\pm$ .015  & .560$\pm$ .008 \\
$\beta_3=\beta_4=0$         & .321$\pm$ .011 & .557$\pm$.002& .373$\pm$ .002  & .559$\pm$ .001 \\
$\beta_4=0$                 & $-$            & .561$\pm$.003& $-$             & .562$\pm$ .006\\
$\beta_i \neq 0$            & .330$\pm$ .020  & .559$\pm$.005& .375$\pm$ .017  & .562$\pm$ .004\\ \bottomrule
\end{tabular}
}
\label{tab:ablation_study}
\end{table}
\begin{figure*}[t]
    \centering
    \includegraphics[width=0.9\linewidth]{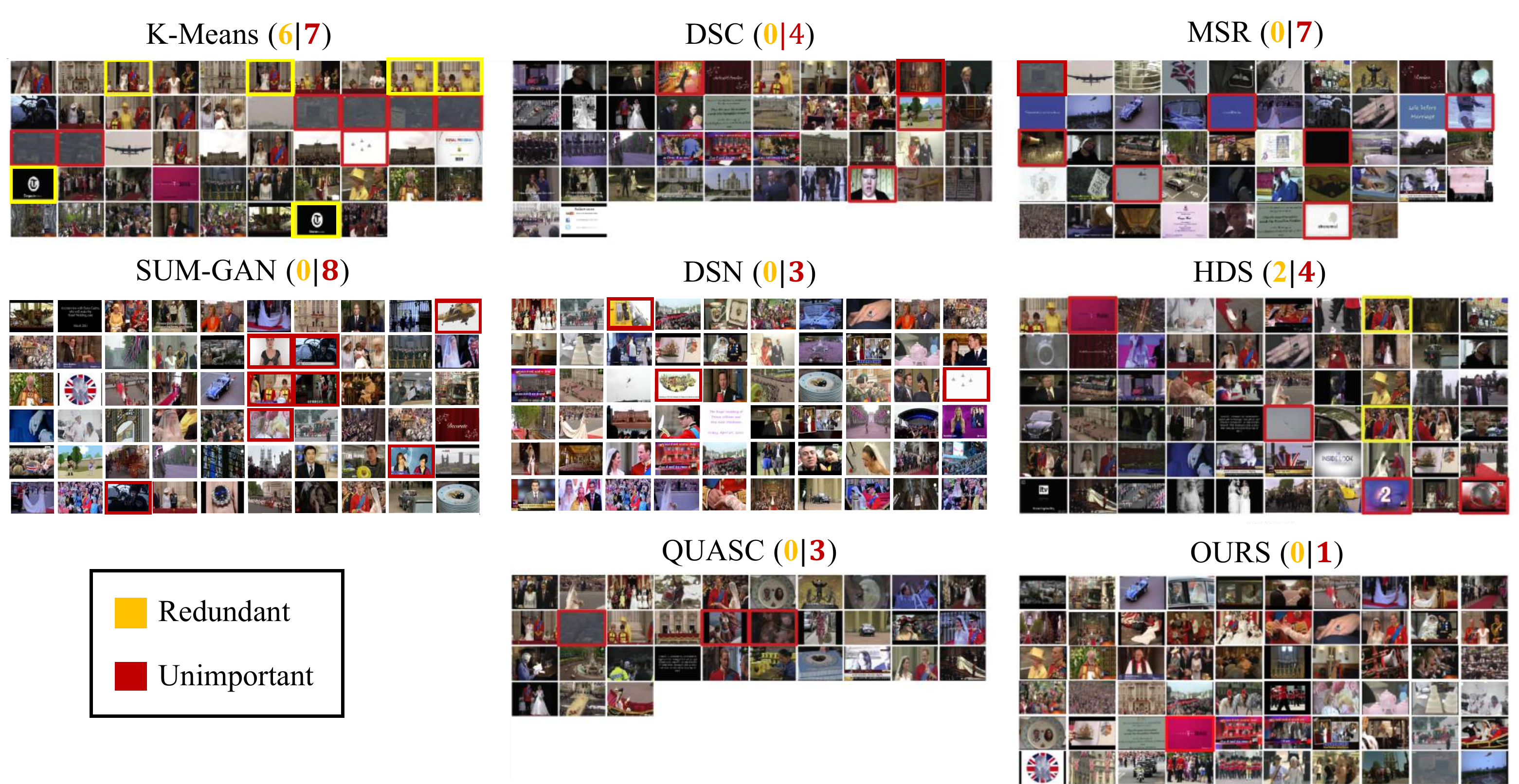}
\caption{Qualitative results for event 1 (Prince William Wedding) by $K$-means~\cite{de2011vsumm},
DSC~\cite{besiris2009combining}, MSR~\cite{besiris2009combining}, QUASC~\cite{Ji2017QueryAwareSC}, MVS-HDS~\cite{Ji2018Hypergraph} and DeepQAMVS, respectively. Frames outlined in red indicate unimportant keyframes, while yellow ones show redundant ones. The number of unimportant and redundant frames are reported on top of every summary.}     
\label{fig:qual_results_2}
\vspace{0.3cm}
\end{figure*}

  \begin{figure*}[t!]
  \hspace{-0.3cm}
    \begin{minipage}{.48\textwidth}
    \resizebox{\columnwidth}{!}{
            \begin{tikzpicture}
            \centering
            \begin{axis}[
                width  = \columnwidth,
                height = 4cm,
                ybar=0pt,
                ymajorgrids, tick align=inside, yticklabel=\pgfmathprintnumber{\tick}\,$\%$,
                ymax=90, 
                tick label style={/pgf/number format/assume math mode=true},
                bar width=4pt,
                ylabel={participants (\%)},
                symbolic x coords={1,2,3,4,5,6,7,8,9,10,AVG},
                xtick=data,
                legend entries={\strut DeepQAMVS, \strut Random, \strut DeepQAMVSwo},
                legend style={
                        at={(0.5,1.3)},
                        font=\small,
                        anchor=north,
                        legend columns=-1,
                        /tikz/every even column/.append style={column sep=0.5cm}
                    },
                legend image code/.code={%
                  \draw[#1] (0cm,-0.1cm) rectangle (0.6cm,0.1cm);
                }  ]
            \addplot[fill=cyan] plot coordinates {(1,57.14) (2,47.62) (3,40) (4,55) (5,10) (6,40) (7,70) (8,55) (9,40) (10,25) (AVG,43.98)};
            \addplot[fill=yellow] plot coordinates {(1,23.81) 
            (2,33.33) (3,25) (4,25) (5,10) (6,35) (7,15) (8,25) (9,30) (10,20) (AVG,24.21)};
            \addplot[fill=gray] plot coordinates {(1,19.05) (2,19.05) (3,35) (4,20) (5,80) (6,25) (7,15) (8,20) (9,30) (10,55) (AVG,31.81)};
            \end{axis}
            \end{tikzpicture}
            }
            \vspace{-0.5cm}
            \caption{\centering \small{Temporal coherence user study for Query IDs ($x$-axis). }}
             \label{fig:user_study}
    \end{minipage}
    \hspace{0.8cm}
    \begin{minipage}{.48\textwidth}
        \resizebox{\columnwidth}{!}{%
        \begin{tikzpicture}
        \centering
        \begin{axis}
        [ 
        width  = \columnwidth,
        height = 3.5cm,
        ybar=0pt,
        bar width=4pt,
        ymajorgrids, tick align=inside,
        ylabel = {Run time (in sec.)},
        symbolic x coords={6,10,4,8,1,9,7,5,3,2,AVG},
        xtick=data,
        yticklabel=\pgfmathprintnumber{\tick},
        tick label style={/pgf/number format/assume math mode=true},
        legend entries={\strut L=30, \strut L=50, \strut L=60},
        legend style={
                        at={(0.5,1.3)},
                        font=\small,
                        anchor=north,
                        legend columns=-1,
                        /tikz/every even column/.append style={column sep=0.5cm}
                    },
        legend image code/.code={%
              \draw[#1] (0cm,-0.1cm) rectangle (0.6cm,0.1cm);
            }   
        ]
        \addplot coordinates {(6,.406) (10,.460) (4,.505) (8,.489) (1,.467) (9,.511) (7,.510) (5,.674) (3,.571) (2,.546) (AVG,.514) };
        \addplot coordinates {(6,.546) (10,.585) (4,.615) (8,.599) (1,.599) (9,.597) (7,.682) (5,.747) (3,.712) (2,.711) (AVG,.639) };
        \addplot coordinates {(6,.615) (10,.628) (4,.687) (8,.643) (1,.616) (9,.725) (7,.724) (5,.756) (3,.744) (2,.744) (AVG,.691) };
        \end{axis}
        \end{tikzpicture}
        }
        \vspace{-0.5cm}
        \caption{\centering  \small Run Time Analysis in seconds. Query IDs ($x$-axis) ordered by total number of input frames.}
            \label{fig:run_time}
    \end{minipage}
\end{figure*}

\subsection{Temporal Coherence User Study}\label{sec:coherence}
Since the provided ground truth summaries are composed of an unordered collection of frames, we resort to a user study to assess the visual coherence of our generated summaries. In total 21 participants are presented with 3 summaries generated from (1) DeepQAMVS, (2) random permutation of the video segments in the DeepQAMVS summary (Random), and (3) DeepQAMVS trained without the temporal coherence reward (DeepQAMVSwo). The participants are asked to select the most coherent summary, paying special attention to transitions between different segments in each video.

From Figure \ref{fig:user_study}, we can see that users preferred our DeepQAMVS summary in 8 out of 10 events.
For events 5 `\textit{Malaysian Airline MH370}' and 10 `\textit{Paris Attack}', users preferred the summaries generated by DeepQAMVSwo. Note that these two events deal with major news incidents and consequently mostly consist of visually similar newscaster segments. In this case, users most likely prefer the resulting summaries from DeepQAMVSwo, as it produces more visually varied summaries due to the higher importance of the diversity reward.

\begin{figure*}[t!]
\subfloat[\centering{\bm{$F_1\!=\!.39$}, $R_{div}\!=\!.68$, $R_{rep}\!=\!.60$, $R_{query}\!=\!.65$, $R_{coh}\!=\!.29$}]{\includegraphics[width=0.9\columnwidth]{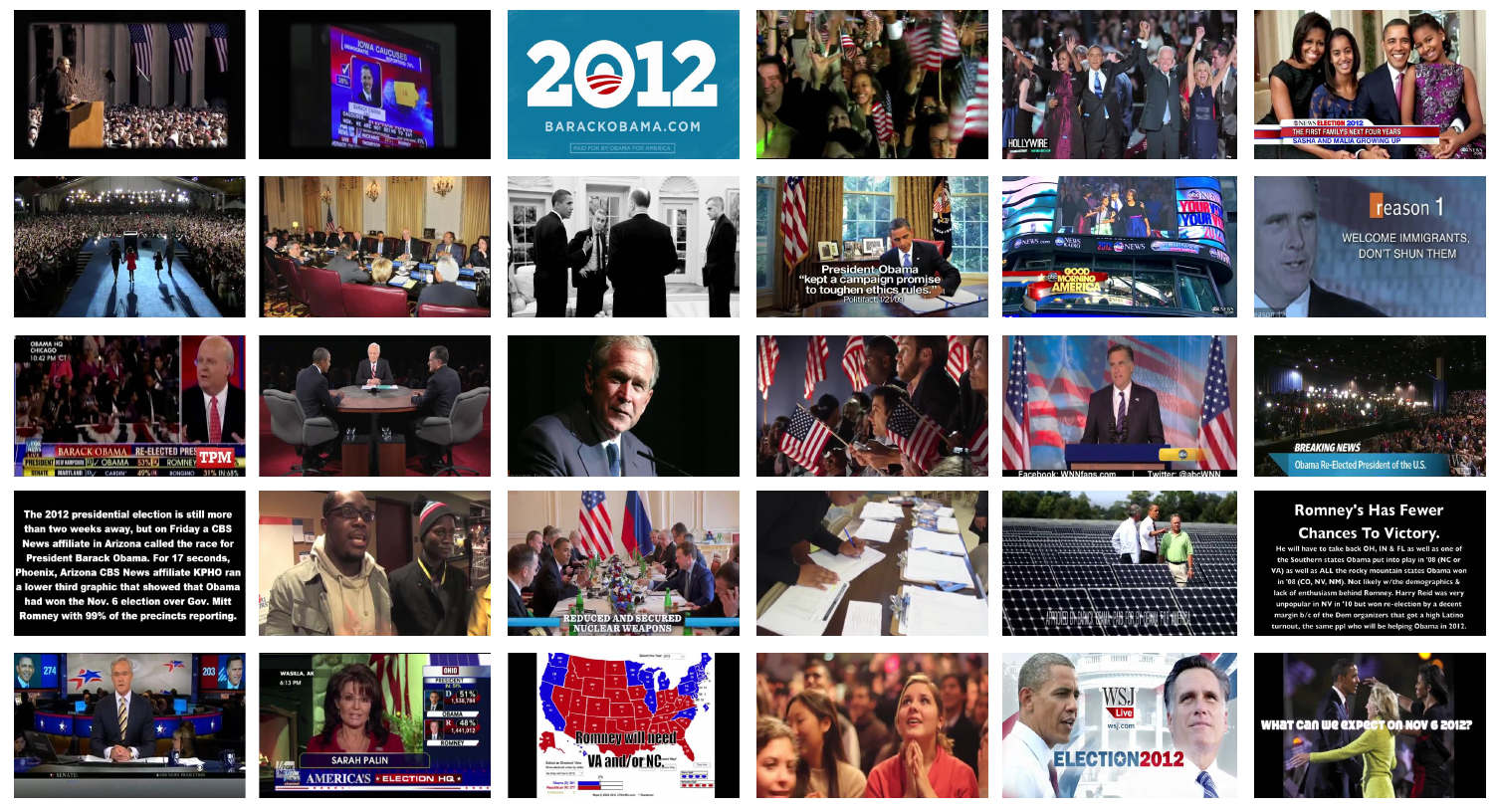}} \hspace{0.5cm}
\subfloat[\centering{\bm{$F_1\!=\!.31$}, $R_{div}\!=\!.67$, $R_{rep}\!=\!.63$, $R_{query}\!=\!.67$, $R_{coh}\!=\!.43$}]{\includegraphics[width=0.9 \columnwidth]{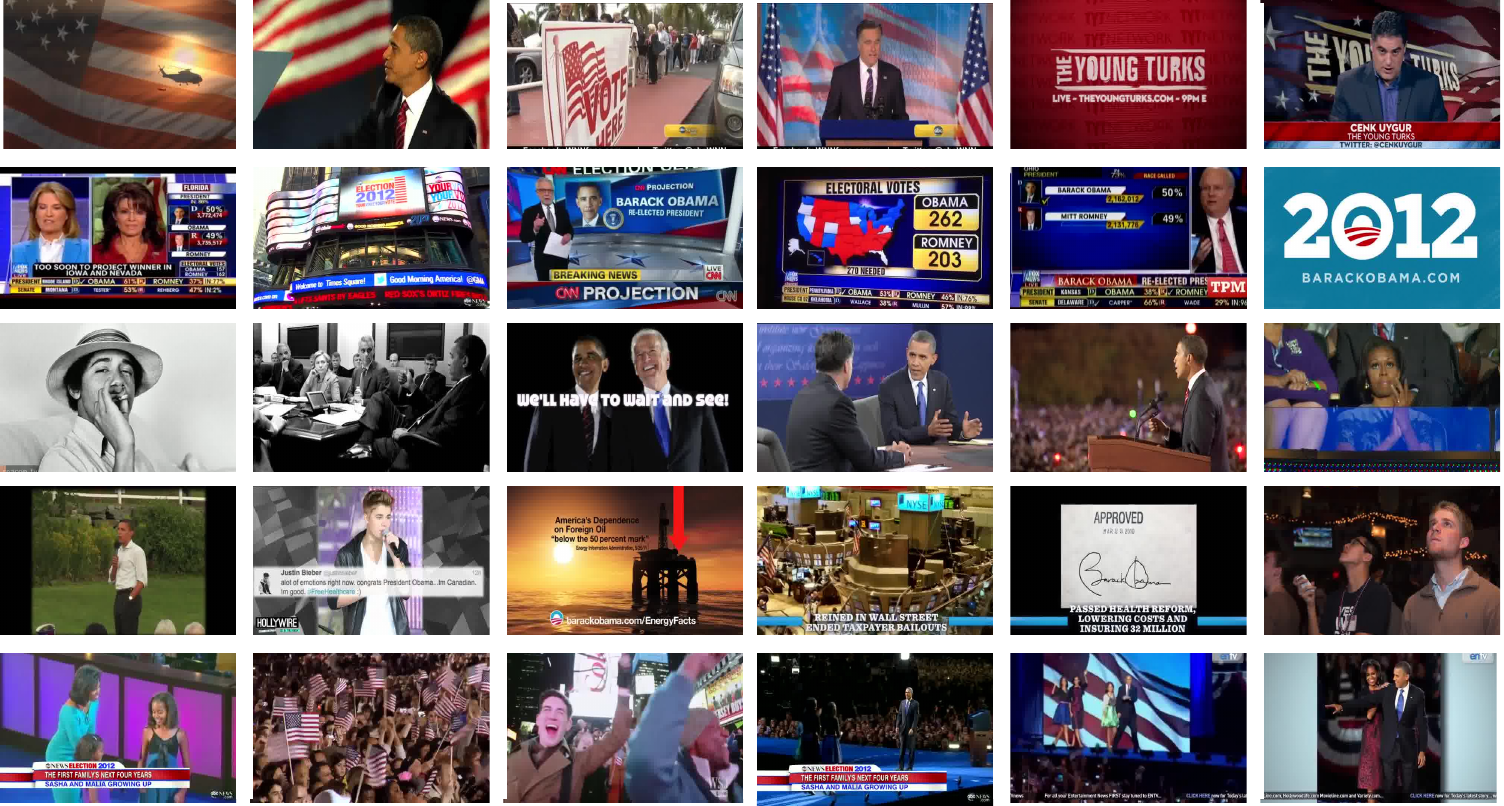}}
\caption{Failure case from event 7 (Obama Re-election). Although, the summary constructed from the ground-truth (left) and the DeepQAMVS generated one (right) are visually and reward-wise comparable, yet there is a remarkable difference in their corresponding F1-scores.}
\label{fig:limitations}
\end{figure*}

\subsection{Qualitative Results}\label{sec:qualitative}
Figure \ref{fig:qual_results_2} illustrates the summaries generated by different methods for the query \textit{Prince William Wedding} (event 1). Visually, we observe that SVS methods choose many irrelevant frames. This is expected as these methods just optimize for diversity and do not take query information into account. QUASC, MWAA and HDS on the other hand have fewer irrelevant frames as they use the web-images to further guide the summarization. Compared to baselines, our method generates summaries with high diversity and selects less unimportant (red bounding box) or redundant frames (yellow bounding box).

\subsection{Run-Time Analysis}\label{sec:runtime} For completeness, we report the run-time of our model in Figure \ref{fig:run_time} for summary lengths $30$, $50$ and $60$. We observe that we scale linearly with the number of input frames and summary length. We do not have access to any QAMVS baseline implementations to measure run-times, but complexity-wise, they all scale polynomially with the number of input frames. 

\subsection{Limitations and Future Work}
Figure \ref{fig:limitations} presents a comparison of two summaries, the ground truth summary (left, (a)) and the summary generated by our DeepQAMVS (right, (b)). While both summaries have high diversity, representativeness and query-adaptability rewards, (b) has a lower F1-score compared to (a). This showcases the limitations of (1) the F1-score as a metric to assess the summary and (2) the subjectivity of the ground truth summaries. 
The F1-score relies solely on the \textit{visual overlap} between the selected frames and the ground truth using pixel-level distances, which are highly sensitive to zooming, shifting and camera angle. 

In fact, Otani \etal \cite{Otani_2019_CVPR} showed that randomly generated summaries achieve comparable or better performance to the state-of-the-art methods when evaluated using the F1-score on two SVS datasets, SumMe \cite{gygli2014creating} and TVSum \cite{song2015tvsum}. Note that the ground truth in their case consists of importance scores associated with every frame. Otani \etal \cite{Otani_2019_CVPR} proposed a new evaluation protocol based on the correlation between the ranking of the estimated scores and the human-annotated ones (Kendall \cite{kendall1945treatment} and Spearman \cite{zwillinger1999crc} correlation coefficients). This metric shows the expected intuition, \ie across human-annotated summaries, the correlation metric is high. In contrast, the correlation between the randomly generated and state-of-the-art summarization methods is small. 

Unfortunately, this metric is not applicable to QAMVS. To see this consider the following: if the ground truth consists of importance scores, redundant frames representing an important event will have high scores across videos. 
Hence, a ranked list of ground truth scores contains redundant frames, which leads to a sub-optimal summary resulting in high Spearman/Kendall scores. 
To fix this, we believe that a metric combining \textit{visual}, \textit{textual} and \textit{temporal order} overlap would lead to a better evaluation protocol. 
Few papers have proposed metrics based on the \textit{textual overlap} in the past. In particular, Yeung \etal \cite{yeung2014videoset} annotated segments in videos with sentences. The ground truth and selected segments are compared using a similarity metric for text summarization (ROUGE). Textual annotation could be very expensive for  QAMVS. However, recent advances in image captioning could be leveraged to automate the process. In this paper, we design a user study to assess the coherence of the produced summaries. However, user-studies are expensive, subjective and not reproducible. Instead, a ranking correlation measure between a list of textual concepts from the ordered ground truth frames and the ones from the proposed summary may serve as a better metric, similar to \cite{Otani_2019_CVPR}. 

Beyond the evaluation metric, training to optimize for the temporal coherence still has room for improvement. Although using the proposed reward results in visually smoother transitions, it did not lead to an overall clear story in the final summary. Embedding frames/web-images in a shared vision-language domain \cite{plummer2017enhancing} could permit to leverage advances in text summarization.  Also, the field could benefit from new benchmarks with more events and shot-level text annotations to enable a wider range of techniques and evaluation metrics.

\section{Conclusion}
In this work, we present DeepQAMVS, the first end-to-end trainable model for query-aware multi-video summarization. DeepQAMVS leverages a pointer network with hierarchical attention to fuse information from video frames, web images and textual meta-data. In addition, we introduce two novel rewards that capture query-adaptability and temporal coherence. Quantitative comparisons with an extensive set of SVS and MVS baselines and thorough qualitative analysis showcase that our model can generate a temporally coherent, query-adaptive, diverse and representative summary from a collection of retrieved videos, achieving state-of-the-art results on the MVS1K dataset. QAMVS needs more community attention and research efforts to tackle the discussed limitations and therefore provide an efficient and robust technology to leverage the exponentially growing online video content.

\section{Acknowledgements}
This work is supported in part by the IBM-Illinois C3SR and by the National Science Foundation under Grant No.\ 1718221, 1801652, 2008387, 2045586.

\bibliographystyle{ACM-Reference-Format}
\bibliography{egbib}

\end{document}